\pdfoutput=1
\documentclass[11pt]{article}

\usepackage[final]{acl}

\usepackage{times}
\usepackage{latexsym}
\usepackage{amsfonts}
\usepackage{amsmath}
\usepackage{subcaption}
\usepackage{graphicx}
\usepackage{scalefnt} %用于设置字体

\usepackage{graphicx}
\usepackage{xcolor}
\usepackage{CJKutf8}
\usepackage{tabularx}
\usepackage{booktabs}
\usepackage{makecell}
\usepackage[misc]{ifsym}
\usepackage{amssymb}% http://ctan.org/pkg/amssymb
\usepackage{pifont}% http://ctan.org/pkg/pifont
\usepackage[export]{adjustbox}
\usepackage{multirow}

\newcommand{\zhsmall}[1]{\begin{CJK*}{UTF8}{gbsn}\small{#1}\end{CJK*}}

\newcommand{\blue}[1]{\textcolor{blue}{#1}}

\newcommand{\orange}[1]{\textcolor{orange}{#1}}

\usepackage[T1]{fontenc}
\usepackage[utf8]{inputenc}

\usepackage{microtype}

\usepackage{inconsolata}

\usepackage{graphicx}

\title{Cross-Lingual Transfer of Cultural Knowledge:\\An Asymmetric Phenomenon}

\author{Chen Zhang,\ \  Zhiyuan Liao,\ \ Yansong Feng\thanks{Corresponding author.} \\
Wangxuan Institute of Computer Technology, Peking University \\
{\tt \{zhangch,fengyansong\}@pku.edu.cn} \\
{\tt liaozy@stu.pku.edu.cn}\\
}

\begin{document}
\maketitle
\begin{abstract}
Despite substantial research efforts evaluating how well large language models~(LLMs) handle global cultural diversity, the mechanisms behind their cultural knowledge acquisition, particularly in multilingual settings, remain unclear.
We study this question by investigating how cultural knowledge transfers across languages during the language adaptation of LLMs, a process where an LLM is continually pre-trained to learn another language.
We introduce an interpretable framework to study this transfer, ensuring training data transparency and controlling transfer effects. 
Through a study of four non-Anglophonic cultures, we observe bidirectional cultural transfer between English and other high-resource languages, while low-resource languages primarily transfer knowledge to English with limited reverse flow.
To explain this asymmetric phenomenon, we propose a frequency-based hypothesis: cultural knowledge appearing more frequently in the pretraining data transfers more easily, which is supported by empirical analysis of the training corpora. 
We hope our findings could inform future research on knowledge transfer and promote the development of culturally aware models, particularly for low-resource languages.
\end{abstract}

\section{Introduction}
Although large language models (LLMs) have made significant progress in processing diverse languages~\cite{huang2024survey,dang2024aya}, they face the challenge of addressing the complexity of global cultural diversity~\cite{hershcovich-etal-2022-challenges,pawar2024survey,liu2024culturally,10.1162/COLI.a.14}. 
Existing research primarily evaluates whether LLMs possess adequate cultural knowledge of non-English-speaking or non-Anglophonic communities~\cite{yin-etal-2022-geomlama,fung2024massively,shi2024culturebank,li2024culturegen}. 
However, the sources and mechanisms by which LLMs acquire cultural knowledge, particularly in multilingual settings, remain largely unexplored.

\begin{figure}[t]
\centering
\includegraphics[width=0.9\columnwidth]{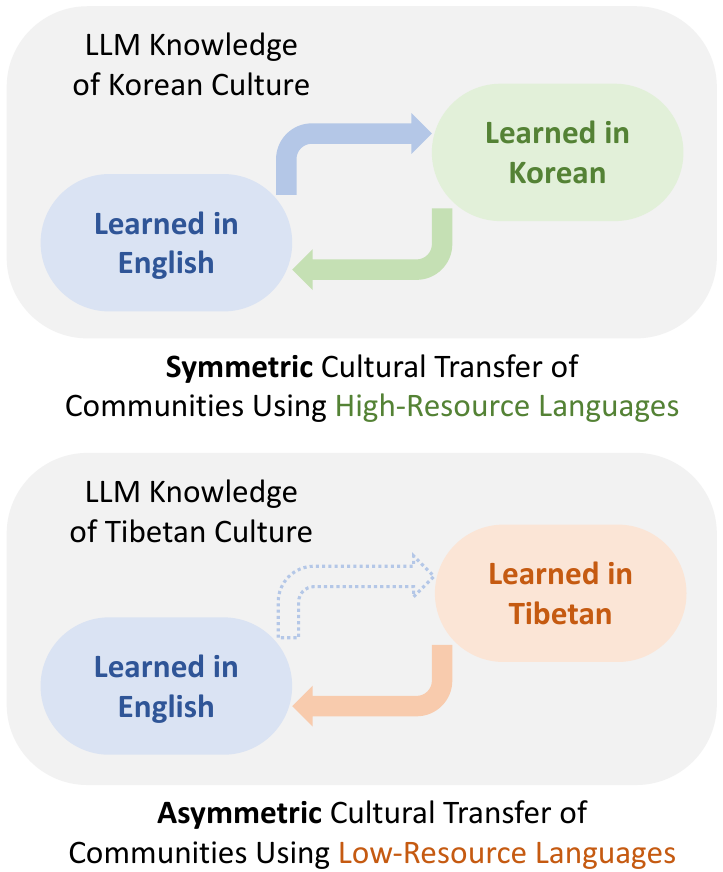}
\caption{LLMs might exhibit different patterns of cross-lingual transfer of cultural knowledge when continually pretrained in non-English languages. The transfer tends to be bidirectional for cultural knowledge of communities using high-resource languages, but often remains asymmetric for low-resource ones.}
\label{fig:first_fig}
\end{figure}

In this paper, we investigate the dynamics of cultural knowledge acquisition in LLMs.
We particularly focus on the mechanisms of cross-lingual transfer, a widely observed phenomenon when LLMs are adapted to new languages~\cite{ye2023language,hu2024large,zhao2024tracing,etxaniz2024bertaqa}.
This investigation faces two main challenges. First, the opacity of LLMs' training data and procedures limits the feasibility of conducting interpretable experiments, making it difficult to analyze the sources and influencing factors of knowledge transfer.
Second, when performance improvements are observed on cultural knowledge questions for a given language, it is challenging to disentangle whether these gains result from improved language proficiency or from cross-lingual knowledge transfer.

\begin{figure*}[t]
\centering
\includegraphics[width=1.9\columnwidth]{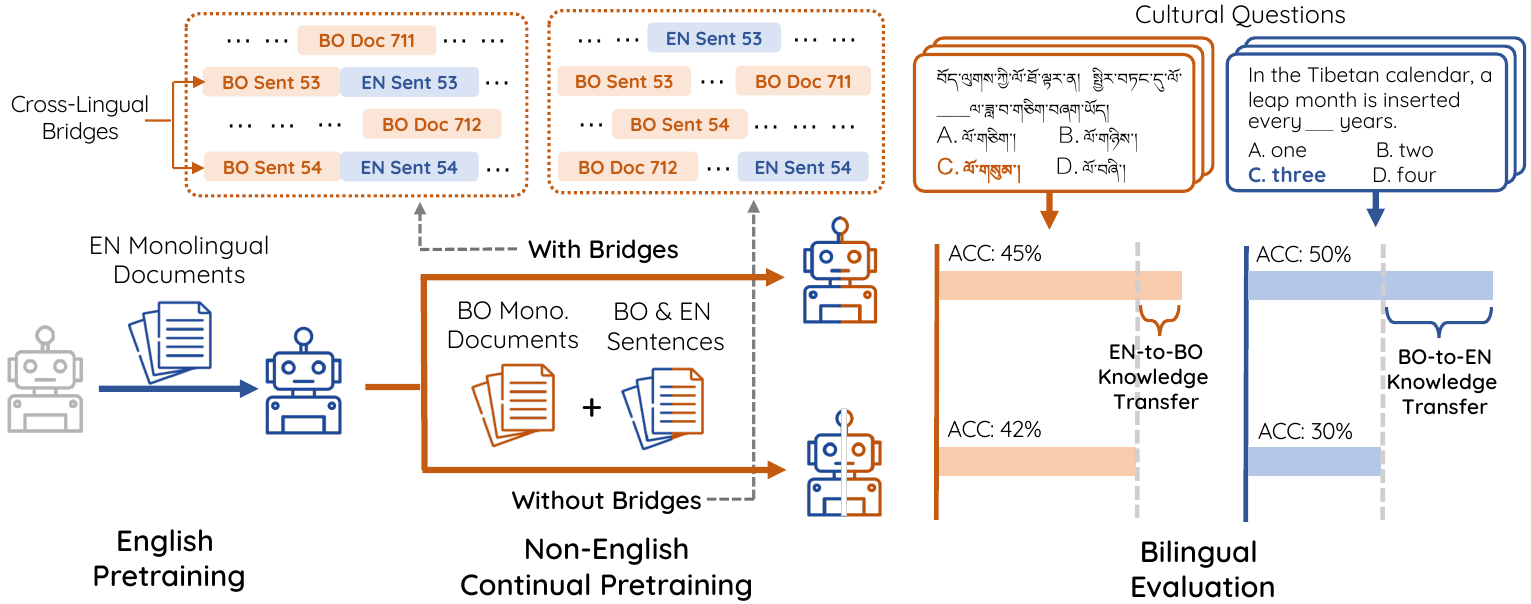}
\caption{Our framework for studying the cross-lingual transfer of cultural knowledge. We use the transfer between English and Tibetan (bo) as an illustrative example.}
\label{fig:pipeline}
\end{figure*}

To address these challenges, we design a new research framework that allows a more controlled and interpretable study on the cross-lingual transfer of cultural knowledge. 
Our framework has three key features.
First, it emphasizes the transparency in training data, making it possible to locate the source of cultural knowledge.
Specifically, we pretrain a base model from scratch using carefully filtered English Wikipedia data, followed by continual pretraining on corpora in other languages.
Second, our framework isolates the effects of cross-lingual transfer from the improved language proficiency.
We employ two continual pretraining settings, one that allows cross-lingual transfer with explicit cross-lingual bridges of parallel co-occurrence and the other that minimizes the chances of such transfer.
The performance gap between these two settings serves as an estimate of cross-lingual transfer effects.
Third, using bilingual parallel probing questions, our framework can analyze the transfer in both directions: from English to non-English languages and vice versa.

We apply this framework to study the cultural transfer of four non-Anglophonic communities: Koreans in South Korea, Han Chinese, Tibetans, and Mongols in China, whose native languages are Korean (ko), Chinese (zh), Tibetan (bo), and Mongolian (mn), respectively.
During continual pretraining on these languages, we observe notable differences in cross-lingual cultural knowledge transfer, as shown in Figure~\ref{fig:first_fig}. 
Specifically, for high-resource languages (Chinese and Korean), knowledge transfer occurs bidirectionally between English and the target languages.
However, for lower-resource languages (Tibetan and Mongolian), we find an asymmetric transfer pattern: knowledge transfer from low-resource languages to English is more pronounced than the reverse direction.

Based on these observations, we hypothesize that cultural knowledge that appears in the training corpus more frequently is more likely to transfer across languages, inspired by previous findings that frequency of appearance during training is an important factor in the monolingual knowledge acquisition of LLMs~\cite{chang2024how,li2025attributing}.
To test this hypothesis, we analyze our training data by estimating the number of occurrences for each cultural knowledge item in both the English and non-English corpora.
Our findings reveal that for low-resource languages, cultural knowledge appears significantly more frequently in non-English corpora than in the English corpus, potentially contributing to the observed asymmetry in transfer. 
An instance-level analysis further shows that cultural knowledge items that transfer across languages tend to have higher-than-average frequencies of occurrence in the corpus, providing additional evidence for our hypothesis.

Our main contributions are as follows:
(1) We introduce a novel framework towards an interpretable study of cross-lingual knowledge transfer, effectively isolating transfer effects from other influencing factors; % 什么叫transfer effects啊
(2) We investigate the dynamics of knowledge transfer for four non-Anglophonic cultural communities, revealing an asymmetric transfer pattern for low-resource languages.
(3) We propose a frequency-based hypothesis to explain the asymmetric transfer phenomenon, supported by empirical analysis over pretraining data.
Our data and code are publicly available to the community\footnote{\url{https://github.com/luciusssss/cross-lingual-culture}}.

\section{Methodology}
\label{sec:method}
As shown in Figure~\ref{fig:pipeline}, our framework consists of three steps: pretraining in English, controlled continual pretraining in non-English languages, and bilingual evaluation on cultural questions. It emphasizes transparent training data, decoupling of transfer effects, and bilingual parallel evaluation.

\paragraph{Transparent Pretrainining}
Instead of using LLMs trained with closed-source training data, we advocate training smaller models from scratch with transparent data.
In this way, we can clearly observe the process of knowledge transfer and trace the sources of learned knowledge from the corpus.
Specifically, we train a 0.5B model from scratch using an English Wikipedia that filters out all non-Latin characters.
Afterwards, we continually pretrain it with corpora in other languages and track how the cultural knowledge transfers throughout the training process.

\paragraph{Decoupling of Transfer Effects in Continual Pretraining}
To disentangle the effects of cross-lingual transfer from improved language proficiency, we systematically control contributing factors, including language similarities, lexical overlaps, and parallel co-occurrences~\cite {radford2019language,blevins-zettlemoyer-2022-language,philippy-etal-2023-towards}.
We design two distinct settings for continual pretraining: one that facilitates cross-lingual transfer and the other that minimizes the chances of transfer as much as possible. 
The performance gap between these two settings can serve as an indicator of the effect of cross-lingual transfer.

Specifically, we select languages using non-Latin writing systems for study, which naturally have little lexical overlap with English.
We further ensure that the English corpus contains no non-Latin characters and the non-English corpora contain no Latin characters, maximizing the isolation between languages during pretraining.

To introduce bridges of cross-lingual transfer to one of the settings, we incorporate parallel sentences into the continual pretraining data, where each pair of parallel sentences is concatenated and mixed with other monolingual training data, as illustrated in Figure~\ref{fig:pipeline}.
This approach has been proven to be effective in helping LLMs learn cross-lingual mappings during pretraining~\cite {anil2023palm,wei2023polylm,lin2024recipe}.
In contrast, the other setting without cross-lingual bridges uses the same training data, but we purposely prevent paired parallel sentences from co-occurring. 
Instead, the two sentences in each parallel pair are treated as independent documents, shuffled within the training data. This eliminates the possibility of learning cross-lingual alignment through bilingual co-occurrence in a shared context.

\paragraph{Bilingual Parallel Evaluation}
For the checkpoints throughout the continual pretraining process, we evaluate them with cultural probing questions in both English and non-English.
% To study the transfer bidirectionally, each cultural question is presented 
When using the non-English-version questions for evaluation, the performance gap between the settings with and without bridges indicates the transfer from English to non-English languages. Similarly, the English-version questions are used to probe the transfer from non-English to English.

\section{Experiments}

\paragraph{Studied Cultures}
We select four cultural communities, Koreans in South Korea, Han Chinese, Tibetans, and Mongols in China, for our study. Two of them are associated with high-resource languages (Korean and Chinese), and the other two with low-resource languages (Tibetan and Mongolian). 
See more information about these communities and their cultures in Appendix~\ref{app:culture_info}.
The languages are deliberately chosen to minimize the cross-lingual transfer with English, as discussed in Section~\ref{sec:method}.

\begin{figure*}[t]
\centering
\includegraphics[width=2.05\columnwidth]{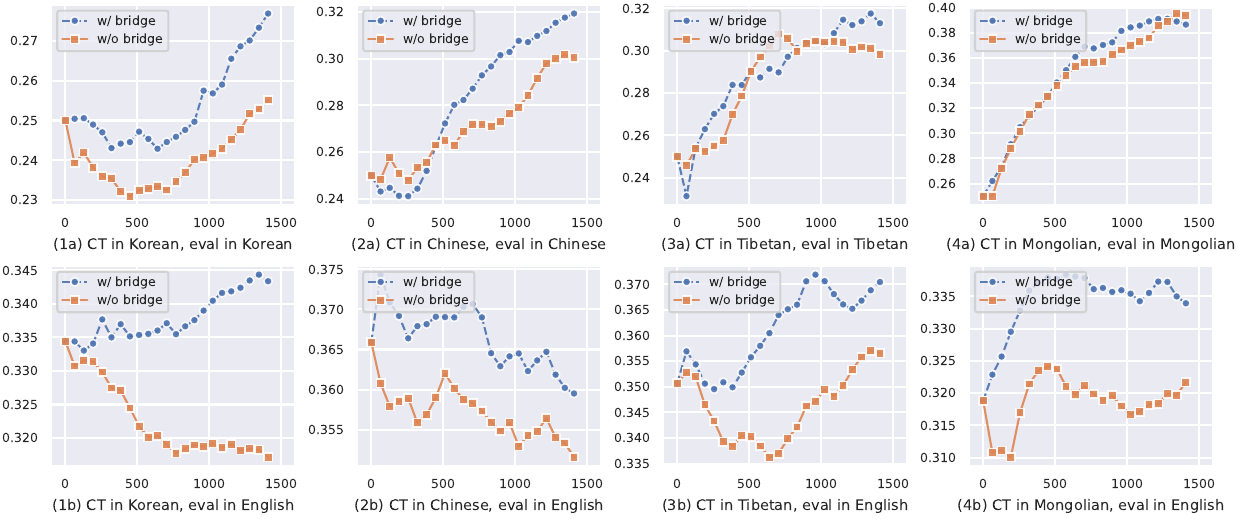}
\caption{
Accuracy on the non-Anglophonic cultural questions with different continual pretraining (CT) steps under different settings. We use EMA smoothing when plotting the figures, with the weight set to 0.8.}
\label{fig:ct_process}
\end{figure*}

\paragraph{Pretraining}
We pretrain a model from scratch with the architecture of Qwen-2.5-0.5B~\cite{yang2024qwen2} using English Wikipedia, amounting to 5B tokens.
We then continually pretrain it for each non-English language for 1,500 steps. The batch size is set to 0.5M tokens.
% The ratio between monolingual and parallel corpora is 1:1.
See details of pretraining corpora and implementation in Appendix~\ref{app:pretraining}.

\paragraph{Evaluation}
Since small models trained from scratch exhibit limited instruction-following capabilities, we use cloze-style questions for evaluation.
For Korean culture, we sample cultural questions from CLIcK~\cite{kim-etal-2024-click} and convert them into cloze-style format using GPT-4o.
For the other three cultures in China, we collect questions from \zhsmall{《中国少数民族》} (\textit{Ethnic Minorities in China}; \citealp{ethic2009}), a comprehensive resource on the history and customs of ethnic minorities in China. 
Specifically, given each paragraph of the book, we ask GPT-4o to generate multi-choice cultural questions, which are then verified and translated by native speakers.
In total, we collect between 276 and 598 questions for each culture. 
See details in Appendix~\ref{appendix:annotation}.

\paragraph{Results and Analyses}
In Figure~\ref{fig:ct_process}, we illustrate how the accuracy of probing questions related to non-Anglophonic cultures evolves throughout the continual pretraining process\footnote{Our main experiments study the transfer of non-Anglophonic cultures. We additionally conduct experiments on the transfer of knowledge exclusively appearing in the Anglophonic culture. See Appendix~\ref{app:anglo_transfer} for details. }.
When the evaluation questions are written in non-English languages (the first row in Figure~\ref{fig:ct_process}), the gaps between the two settings, one with bridges (\blue{blue} lines with circular markers) and the other without (\orange{orange} lines with square markers), can be an estimation of knowledge transferred from English, as opposed to knowledge acquired through continual pretraining.
For relatively high-resource languages (Korean and Chinese), we observe notable English-to-non-English transfer (1a and 2a in Figure~\ref{fig:ct_process}), while for low-resource languages (Tibetan and Mongolian), such transfer is less evident (3a and 4a in Figure~\ref{fig:ct_process}).

Regarding the transfer from non-English to English (the second row in Figure~\ref{fig:ct_process}), we observe clear performance gaps between the settings with and without bridges for all four cultures, indicating that the knowledge of non-Anglophonic cultures flows into English. 
Notably, for most languages (1b, 3b, and 4b in Figure~\ref{fig:ct_process}), this transfer appears to make up for the forgetting of English abilities due to non-English continual pretraining: the performance on English-version questions in the with-bridge settings  continues to improve.

\section{Inspection into Training Data}
\label{sec:hypotheis}
We hypothesize that the cross-lingual transfer of cultural knowledge is largely influenced by its frequency of occurrence in the training corpus, which has been identified as an important factor in monolingual knowledge acquisition~\cite{chang2024how,li2025attributing}.
Specifically, cultural knowledge that appears more frequently in the training corpus of one language is more likely to be transferred to another language. 
We test this hypothesis by estimating the number of occurrences within our training data for each cultural knowledge piece.

\paragraph{Setup} 
For each cultural probing question, we first retrieve the 50 most relevant documents from the corpus. 
We then employ Llama-3.1-70B~\cite{dubey2024llama} to assess whether each retrieved document entails the cultural knowledge in the question. 
Given the substantial differences in corpus sizes between English and non-English languages, we introduce the metric of \textbf{cultural density}, normalizing the count of knowledge appearance by the total number of documents in the corpus.
See Appendix~\ref{app:retrieval} for implementation details.

\begin{table}[t]
\small
\centering
\begin{tabular}{l|cc}
\toprule
 & \textbf{EN Corp.} & \textbf{Non-EN Corp.} \\
\midrule
Koreans & 2.86e-7	 & 5.21e-7 \\
Han Chinese & 2.97e-7	& 2.84e-7 \\
Tibetans & 1.49e-7 & 9.19e-6 \\
Mongols in China& 1.55e-7 &	3.72e-6 \\
\bottomrule
\end{tabular}
\caption{The cultural densities of different cultures in the English and non-English corpora.}
\label{tab:density}
\end{table}

\paragraph{Results}
As shown in Table~\ref{tab:density}, for cultures associated with low-resource languages (Tibetan and Mongolian), their densities in non-English corpora are one order of magnitude higher than those in the English corpus. Thus, the knowledge transfer from non-English to English is more evident than that in the reverse direction.
Meanwhile, for the cultures of higher-resource languages (Korean and Chinese), the cultural densities between English and non-English corpora are in the same order of magnitude. Accordingly, the extent of cross-lingual transfer is similar for both transfer directions. 

Beyond this culture-level comparison, we look at individual data points, examining the subset of cultural knowledge that is successfully transferred between English and non-English (see Appendix~\ref{app:criterion} for selection criteria). 
We find that these transferred knowledge pieces occur more frequently in the corpus than average. 
For cultural knowledge transferred from English to non-English, the average occurrence in the English corpus is 9.0, significantly higher than the overall average of 4.2. 
Similarly, for cultural knowledge transferred from non-English to English, the average occurrence in the non-English corpus is 4.7, compared to the general average of 2.2.
These findings further support our frequency-based hypothesis.

\section{Conclusion}
We investigate cross-lingual cultural knowledge transfer during LLM language adaptation. 
With an interpretable research framework, we observe bidirectional transfer for high-resource languages but an asymmetric pattern for low-resource languages. 
We propose a frequency-based hypothesis to explain this phenomenon, with evidence from the analysis of pretraining corpora. 
We hope that our study could inspire more efforts to uncover the dynamics of knowledge transfer and improve the cultural awareness of LLMs, especially for low-resource languages.

\section*{Limitations}
\paragraph{Model Scale} 
We only use 0.5B models for experiments because our experiments involve pretraining for 16 different settings, which requires considerable computational resources. 
We select our model size as a compromise between the informativity of results and computational cost.

\paragraph{Coverage of Cultures}
To control the factors of cross-lingual transfer in the experiments, we need to use non-Indo-European languages employing non-Latin writing systems, which exhibit greater typological divergence from English and have little lexical overlap with English. This strict requirement significantly narrows the range of suitable cultures for study.
Additionally, the collection, verification, and translation of cloze-style cultural questions demand substantial human effort, as existing cultural evaluation datasets are typically monolingual and open-ended, which are not suitable for our study.

Considering these constraints, we select four culturally representative cases, ensuring a balance between well-represented and underrepresented cultures for a more comprehensive analysis. 
Notably, for the two underrepresented cultural communities, Tibetans and Mongols in China, no existing NLP studies on cultural analysis have considered them. 
Our evaluation dataset thus provides a valuable resource for investigating these low-resource languages and their cultural representations.

\paragraph{Factors of Cross-Lingual Transfer}
In our research framework, we try our best to control the factors of cross-lingual transfer for explainable experiments, including language similarities, lexical overlaps, and parallel co-occurrences, which have been identified as primary contributors to cross-lingual transfer~\cite{radford2019language,blevins-zettlemoyer-2022-language,philippy-etal-2023-towards}. 
Although several studies find that model architectures~\cite{dufter-schutze-2020-identifying} and pretraining objectives~\cite{liu2020study} are related to cross-lingual transfer, their effects have been extensively validated at scale. Also, these model-related variables remain constant across different settings in our experiments, minimizing their influence on our findings.

\paragraph{Retrieval Systems}
Our analysis of the training data involves retrieving documents relevant to the studied knowledge pieces, a common practice in research on knowledge acquisition~\cite{kandpal2023large,li2025attributing,wang2025generalization}.
However, retrieval systems are not guaranteed to be perfect. To mitigate potential inaccuracies, we carefully design and optimize retrieval systems for each language, ensuring robust performance and reliable results. See details in Appendix~\ref{app:retrieval}.

\section*{Acknowledgements}
This work is supported in part by NSFC~(62161160339) and Beijing Science and Technology Program~(Z231100007423011). We thank Xiao Liu, Mingxu Tao, and the anonymous reviewers for their valuable feedback. 
We also thank all the annotators who contributed to the dataset construction.
For any correspondence, please contact Yansong Feng.

\section*{Ethical Considerations}
A potential concern when collecting cultural questions from existing sources, such as online corpora or books, is the risk of misrepresenting cultures, particularly for underrepresented cultures using low-resource languages. 
To address this issue, we ensure that all cultural probing questions are verified by native speakers from the respective communities. This process guarantees that our dataset contains no inaccurate or offensive descriptions of their cultures.
We discuss this process in detail in Appendix~\ref{appendix:annotation}.

\bibliography{anthology,custom}
% \bibliography{custom}

\clearpage
\appendix

\section{Information of Studied Cultures}
\label{app:culture_info}
In Table~\ref{tab:cultural_communities_info}, we briefly introduce the four communities studied in this work, including regions, populations, languages, and writing systems.

\begin{table*}[t]
\small
\centering
\begin{tabular}{l|cc|ccc}
\toprule
 & \textbf{Region} & \textbf{Population} & \textbf{Native Language} & \textbf{ISO 639-1} & \textbf{Writing System} \\
\midrule
Koreans & South Korea & 52M & Korean & ko & Hangul \\
Han Chinese & China & 1.3B & Chinese & zh & Chinese characters \\
Tibetans & China & 7.1M & Tibetan & bo & Tibetan script \\
Mongols in China & China & 6.3M & Mongolian & mn & Traditional Mongolian script \\
\bottomrule
\end{tabular}
\caption{Information of the cultural communities and their languages studied in our paper. }
\label{tab:cultural_communities_info}
\end{table*}

\section{Collection of Cultural Questions}
\label{appendix:annotation}
\paragraph{Collection of Questions for Koreans}
We use the data from CLIcK~\cite{kim-etal-2024-click}, a benchmark evaluating LLMs' knowledge of the Korean language and culture. 
Each data point in it is a multi-choice question.

From the \textit{culture} subset of CLIcK, we select the data points where the question length does not exceed 300 characters and the choice length does not exceed 30 characters. In this way, we filter out the lengthy questions requiring reading comprehension, which are not suitable for the evaluation of small models such as the 0.5B ones in our experiments.

As the questions in CLIcK end with a question mark, we convert them into cloze-style questions with GPT-4o. A cloze-style question ends with a period and has a blank in itself, where the correct answer fits, similar to the format of MLAMA~\cite{kassner-etal-2021-multilingual}.
We further filter the questions that do not meet this requirement after conversion. 
These questions are further translated from Korean to English by GPT-4o. We manually verify the translation results.
Our use of CLIcK is consistent with its intended use.

\paragraph{Collection of Questions for Han Chinese, Tibetans, and Mongols in China}
For the other three cultures in China, we construct probing questions from scratch, as there is no suitable data available.

We excerpt the chapter describing each group from the book \zhsmall{《中国少数民族》}~(\textit{Ethnic Minorities in China}; \citealp{ethic2009}), a book written in Chinese describing the history and custom of China's ethnic minorities.
We then ask GPT-4o to generate multi-choice cultural questions based on each paragraph of the chapter. 
To facilitate the generation of distracting answers to the questions, we additionally provide GPT-4o with paragraphs that discuss the same topic but describe other ethical groups. 

After generation, we first ask native speakers from each cultural community to check whether the questions faithfully reflect their cultures. 
For Chinese and Tibetan, all the questions pass the check, while for Mongolian, 2\% of the generated questions are marked as misrepresenting their cultures.

The collected questions are translated from Chinese into English by GPT-4o.  
We manually verify the translation results.
For the questions related to Tibetan and Mongolian cultures, we ask native speakers to translate them from Chinese into Tibetan and Mongolian, respectively.

The annotators are recruited from universities, who are native speakers of the low-resource languages and proficient in Chinese. 
They are informed of how the collected data will be used.
They are paid approximately \$1 for translating a question, which is fair given the participants’ demographic.

\paragraph{Data Statistics}
In total, we collect 598 questions for Koreans, 328 questions for Han Chinese, 268 questions for Tibetans, and 552 questions for Mongols in China.
See other statistics in Table~\ref{tab:cultural_question_statistics}.

\begin{table}[t]
\small
\centering
\setlength\tabcolsep{2.5pt}
\begin{tabular}{l|rrrr}
\toprule
& \textbf{ko} & \textbf{zh} & \textbf{bo} & \textbf{mn} \\
\midrule
Number of Questions & 598 & 328 & 268 & 276 \\
Avg. Len. of EN Questions & 16.3 & 22.8 & 18.0 & 19.7 \\
Avg. Len. of EN Options & 4.1 & 3.6 & 3.0 & 3.2 \\
Avg. Len. of Non-EN Questions & 20.5 & 17.3 & 91.3 & 111.6 \\
Avg. Len. of Non-EN Options & 5.1 & 2.6 & 14.6 & 20.5 \\
\bottomrule
\end{tabular}
\caption{Statistics of the cultural probing questions. The average lengths are measured by the tokens produced by the tokenizer of Qwen-2.5-0.5B.}
\label{tab:cultural_question_statistics}
\end{table}

\section{Implementation Details}

\subsection{Continual Pretraining}
\label{app:pretraining}
\paragraph{Data}

Regarding English pretraining, we use the whole Wikipedia\footnote{\url{https://huggingface.co/datasets/wikimedia/wikipedia}} for pretraining, which amounts to approximately 5B tokens. We remove all the non-Latin characters in it, ensuring that during pretraining, the model do not see contents written in the four non-English languages of our study.

For the continual pretraining of four non-English languages, we use both monolingual and parallel corpora. Their ratio is 1:1.
Regarding monolingual pretraining corpora, we use Wikipedia for English, Chinese, and Korean. 
For Mongolian and Tibetan, we use MC$^2$~\cite{zhang-etal-2024-mc2}, as the sizes of Wikipedia for these languages are too small for pretraining. 
Regarding parallel corpora, we use the data from Lego-MT~\cite{yuan-etal-2023-lego} for Chinese and Korean. 
For Mongolian and Tibetan, we use the parallel data from the National Basic Science Data Center of China\footnote{\url{https://cstr.cn/16666.11.nbsdc.vtfshbjs}}.
For the setting with cross-lingual transfer bridges, we use the templates to concatenate each pair of parallel sentences into a document. For example, the template for concatenating English-Chinese parallel sentences is \texttt{English: \{english\_setentece\} Chinese: \{chinese\_sentence\}}.

\paragraph{Hyperparameters}
We use Deepseed\footnote{\url{https://github.com/microsoft/DeepSpeed}} for training. 
The model is trained for up to 1500 steps using a batch size of 0.5M tokens, a learning rate of 1e-4, and a warmup ratio of 0.01 on eight A100 GPUs. 
A training step takes approximately 18 seconds.

\subsection{Evaluation}
\label{app:Evaluation}
The probing questions are in a cloze style, each paired with four candidate answers.
We put each answer candidate into the blank in the question and calculate the perplexity of the sentence. The candidate with the lowest perplexity is chosen as the final prediction.
We use the evaluation scripts from \citet{qi-etal-2023-cross}.

\subsection{Knowledge Retrieval}
\label{app:retrieval}
We adopt different methods for the knowledge retrieval of high-pressure and low-resource languages, due to their differences in the availability of mature retrieval techniques. 

\paragraph{Retrieval for High-Resource Languages}
For high-resource languages such as English, Chinese, and Korean, we adopt competitive dense retrieval models that are available for these languages.
We use \texttt{BAAI/bge-small-en-v1.5}\footnote{\url{https://huggingface.co/BAAI/bge-small-en-v1.5}}~\cite{bge_embedding} for English, \texttt{BAAI/bge-small-zh-v1.5}\footnote{\url{https://huggingface.co/BAAI/bge-small-zh-v1.5}} for Chinese, and \texttt{upskyy/bge-m3-korean}\footnote{\url{https://huggingface.co/upskyy/bge-m3-korean}}~\cite{bge-m3} for Korean.
Before retrieval, the documents in the corpora are split into chunks no longer than 5,000 characters.
After retrieval, we use Llama-3.1-70B~\cite{dubey2024llama} to determine whether a retrieved document can entail each query.

\paragraph{Retrieval for Low-Resource Languages}
For low-resource languages, i.e. Tibetan and Mongolian, no dense retrieval models are available, and current LLMs perform poorly on the NLI task in these languages.
Therefore, we mainly adopt lexical-based retrieval and translate queries and documents into high-resource languages for NLI.

Specifically, we first retrieve the top 50 documents using BM25~\cite{robertson2009probabilistic} method, a language-agnostic retrieval algorithm. Before retrieval, the documents are split into chunks no longer than 5,000 characters.
Then, we translate both queries and retrieved documents into Chinese using \texttt{NLLB-200-3.3B}\footnote{The model is further finetuned with Chinese-Tibetan and Chinese-Mongolian parallel data to enhance its translation capabilities. }~\cite{nllb-24}, 
We then use Llama-3.1-70B~\cite{dubey2024llama} to determine the entailment relationship between each query and its 50 retrieved documents.

\subsection{Criterion of Successfully Transferred Instances}
\label{app:criterion}
In our instance-level analysis, we adopt a strict criterion to identify the knowledge pieces successfully transferred across languages.

For a knowledge piece successfully transferred from non-English to English, its corresponding cultural question should (1) be incorrectly answered in the English version before continual pretraining (CT), (2) be correctly answered in the non-English version by the last three CT checkpoints of the no-bridge setting, and (3) be correctly answered in the English version by the last three CT checkpoints of the bridged setting.

For a knowledge piece successfully transferred from English to non-English, its corresponding cultural question should (1) be correctly answered in the English version before CT, (2) be incorrectly answered in the non-English version by the last three CT checkpoints of the no-bridge setting, and (3) be correctly answered in the non-English version consistently by the last three CT checkpoints of the bridged setting.

\section{Transfer of Knowledge from Anglophonic Cultures}
\label{app:anglo_transfer}

\paragraph{Data Collection}
% to be revised 
It is untrivial to construct cultural questions for \textit{English culture} as English is the native language of many countries or communities, whose cultures may differ greatly.
As a workaround, we study the information of celebrities from the core Anglosphere (Australia, Canada, New Zealand, the United Kingdom, and the United States).
This kind of information is more likely to appear in the English corpus than in non-English corpora.

Specifically, we sample 1,000 persons from Wikidata who meet the following requirements: (1) their nationality is in the core Anglosphere, (2) they have only English Wikipedia pages and do not have pages in the four languages of our study (Chinese, Korean, Mongolian, and Tibetan). 
We then obtain their attributes, including nationality, date of birth,  place of birth, occupation, and educational background.
We further ensure that their names and attributes only appear in the English corpus, and not in the corpus of other languages.

These person-attribute pairs are then converted to multi-choice questions using templates in different languages, such as \texttt{\{person\} was born in \{birthplace\}} and \texttt{\{person\}}\zhsmall{出生在}\texttt{\{birthplace\}}. 
We directly use the person names and their attributes in their original English forms, as translating them requires substantial human labor.
In total, we collect 3,073 such questions.

\paragraph{Experiments and Results}
We investigate whether Anglophonic cultural knowledge acquired during English pretraining can be transferred to other languages after continual pretraining.
Following the experimental setup outlined in Section~\ref{sec:method}, we compare two settings: one with cross-lingual transfer bridges and one without. After 1,500 steps of continual pretraining, we evaluate the models using non-English versions of Anglophonic cultural questions.

Our results show that the setting with transfer bridges achieves, on average, 20\% higher accuracy compared to the setting without bridges, indicating that Anglophonic cultural knowledge can be transferred to non-English languages.
This finding is also in line with our frequency-based hypothesis in Section~\ref{sec:hypotheis}: Anglophonic cultural knowledge appears more frequently in English pretraining data than in non-English data, making it more easily transferable to other languages.

\end{document}